%% file: main.tex
\documentclass{article}

\PassOptionsToPackage{numbers, compress}{natbib}
\usepackage[preprint]{neurips_2025}

\usepackage[utf8]{inputenc} % allow utf-8 input
\usepackage[T1]{fontenc}    % use 8-bit T1 fonts
\usepackage{hyperref}       % hyperlinks
\usepackage{url}            % simple URL typesetting
\usepackage{booktabs}       % professional-quality tables
\usepackage[table]{xcolor} % for table row colors
\usepackage{amsfonts}       % blackboard math symbols
\usepackage{nicefrac}       % compact symbols for 1/2, etc.
\usepackage{microtype}      % microtypography
\usepackage{colortbl}

\usepackage{graphicx}
\usepackage{multirow}
\usepackage{xspace}
\usepackage{enumitem}
\usepackage[most]{tcolorbox}
\usepackage{fontawesome} % for icons
\usepackage{tikz}
\usepackage{wrapfig}
\usetikzlibrary{shadows}
\usepackage{tabularx,adjustbox}
\usepackage{subcaption}
\usepackage{tabularx}

\newtcbox{\codeblock}{
  colback=gray!10,
  colframe=black!20,
  left=0pt,right=0pt,top=0pt,bottom=0pt,
  boxrule=0.2pt,
  fontupper=\footnotesize\ttfamily,
  sharp corners,
  nobeforeafter,
  width=\linewidth,
  halign=flush left,
}
\definecolor{cream}{RGB}{255, 251, 234}
\definecolor{goldenstar}{RGB}{255, 204, 0}
\definecolor{midnightgreen}{rgb}{0.0, 0.29, 0.33}
\definecolor{deepgreen}{HTML}{0aa344}
\definecolor{deeppurple}{HTML}{7030a0}
\definecolor{deepblue}{HTML}{171d91}
\definecolor{brown}{HTML}{843c0c}
\definecolor{shadered}{HTML}{ffe5e5}
\definecolor{shadegreen}{HTML}{e5f7ed}
\definecolor{msftBlack}{RGB}{0,0,0}
\definecolor{lightred}{RGB}{255,163,163}
\definecolor{deepred}{RGB}{146,0,0}
\definecolor{headergray}{RGB}{240,240,240}
\definecolor{groupblue}{RGB}{220,235,250}
\definecolor{groupgreen}{RGB}{220,250,220}
\definecolor{grouppink}{RGB}{250,230,240}
\definecolor{groupyellow}{RGB}{255,250,205}
\definecolor{groupgray}{RGB}{245,245,245}

\tcbset{
  takeawaystyle/.style={
    colback=cream,
    colframe=black!30,
    boxrule=0.4pt,
    arc=3pt,
    left=6pt,
    right=6pt,
    top=6pt,
    bottom=6pt,
    fonttitle=\bfseries,
    coltitle=black,
    enhanced,
    drop shadow={shadow xshift=1pt, shadow yshift=-1pt, opacity=0.15},
  }
}

\title{MIRIX: Multi-Agent Memory System for\\LLM-Based Agents}

\newif\ifshowcomments
\showcommentstrue

\newcommand{\ours}{MIRIX\xspace}

\author{%
  Yu Wang, Xi Chen \\
  MIRIX AI \\
  \texttt{yuw164@ucsd.edu}\\
  \texttt{xc13@stern.nyu.edu} \\
\url{https://mirix.io/}
}

\begin{document}

\maketitle

\begin{abstract}
Although memory capabilities of AI agents are gaining increasing attention, existing solutions remain fundamentally limited. Most rely on flat, narrowly scoped memory components, constraining their ability to personalize, abstract, and reliably recall user-specific information over time. To this end, we introduce MIRIX, a modular, multi-agent memory system that redefines the future of AI memory by solving the field’s most critical challenge: enabling language models to \emph{truly} remember. Unlike prior approaches, MIRIX transcends text to embrace rich visual and multimodal experiences, making memory genuinely useful in real-world scenarios. MIRIX consists of six distinct, carefully structured memory types: Core, Episodic, Semantic, Procedural, Resource Memory, and Knowledge Vault, coupled with a multi-agent framework that dynamically controls and coordinates updates and retrieval. This design enables agents to persist, reason over, and accurately retrieve diverse, long-term user data at scale.
We validate MIRIX in two demanding settings. First, on ScreenshotVQA, a challenging multimodal benchmark comprising nearly 20,000 high-resolution computer screenshots per sequence, requiring deep contextual understanding and where no existing memory systems can be applied, MIRIX achieves 35\% higher accuracy than the RAG baseline while reducing storage requirements by 99.9\%. Second, on LOCOMO, a long-form conversation benchmark with single-modal textual input, MIRIX attains state-of-the-art performance of 85.4\%, far surpassing existing baselines. These results show that MIRIX sets a new performance standard for memory-augmented LLM Agents. To allow users to experience our memory system, we provide a packaged application powered by \ours. It monitors the screen in real time, builds a personalized memory base, and offers intuitive visualization and secure local storage to ensure privacy.
\end{abstract}

\input{1_introduction}

\input{9_usecase}

\input{3_method}
\input{4_experiments}

\input{2_related_work}
\input{5_conclusion}

\clearpage
\bibliographystyle{plainnat}
\bibliography{ref}

\clearpage

\appendix
\input{6_appendix}

\end{document}

%% file: 1_introduction.tex
\section{Introduction}
Recent advancements in large language model (LLM) agents have focused primarily on improving their capabilities in complex task execution—ranging from code debugging and repository management to autonomous web browsing. While these functionalities are crucial, another foundational yet underexplored dimension is memory: the ability of agents to persist, retrieve, and utilize past user-specific information over time. Human cognition relies heavily on memory—recalling conversations, recognizing patterns, and adapting behavior based on prior experience. Analogously, memory mechanisms in LLM agents are essential for delivering consistent, personalized interactions, learning from feedback, and avoiding repetitive queries.
However, most LLM-based personal assistants remain stateless beyond their current prompt window, retaining no lasting memory unless context is explicitly re-provided. This limitation hinders their long-term usability, especially in real-world settings where users expect assistants to evolve, recall, and personalize over time \cite{lscs}.

To address this, a range of memory-augmented systems have been proposed. One common approach is the use of knowledge graphs, as seen in systems like Zep~\citep{zep} and Cognee~\citep{cognee}. These frameworks are well-suited for representing structured relationships between entities but struggle to model sequential events, emotional states, full-length documents, or multi-modal inputs such as images. Another approach involves flattened memory architectures that store and retrieve textual chunks using vector databases. Examples include Letta~\citep{memgpt}, Mem0~\citep{mem0}, and ChatGPT’s memory system. Letta divides its memory into components—recall memory for conversation history, core memory for preferences, and archival memory for long documents—while ChatGPT focuses primarily on core and recall memories. Mem0 adopts a memory system that contains flattened facts distilled from the user inputs, which serves similar roles as Letta's archival memory while being more distilled. While prevalent, these memory systems face several challenges: (1) Lack of compositional memory structure: Most approaches store all historical data in a single flat store without routing into specialized memory types (e.g., procedural, episodic, semantic), making retrieval inefficient and less accurate. (2) Poor multi-modal support: Text-centric memory mechanisms fail when the majority of the input is non-verbal (e.g., images, interface layouts, maps). (3) Scalability and abstraction: Storing raw inputs, especially images, leads to prohibitive memory requirements, with no effective abstraction layer to summarize and retain only salient information. 

\begin{figure*}
    \centering
    \includegraphics[width=\linewidth]{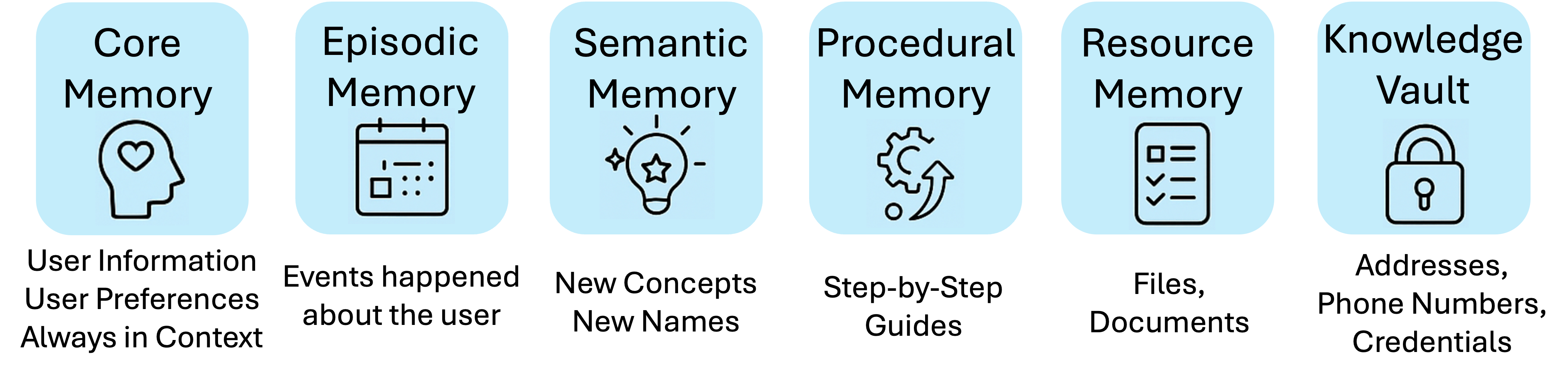}
    \caption{The six memory components of \ours, each providing specialized functionality.}
    \label{fig:overall_structure}
\end{figure*}

To address the limitations of existing memory systems, we argue that effective \textbf{Routing} and \textbf{Retrieving} are the key capabilities a memory-augmented agent must possess. Most current systems focus primarily on Short-Term and Long-Term Memory~\citep{memgpt,caim,cognitive_memory_in_llms}, with some incorporating Mid-Term Memory~\citep{memoryos}. In contrast, we draw inspiration from works that explore more specialized memory types, including episodic memory~\citep{episodic_memory_missing_piece,echo,arigraph}, semantic memory~\citep{memory_consciousness_and_llm}, and procedural memory~\citep{procedural_memory_all_you_need}. Building on these foundations, we propose a more comprehensive architecture consisting of six memory components: \textbf{Core Memory}, \textbf{Episodic Memory}, \textbf{Semantic Memory}, \textbf{Procedural Memory}, \textbf{Resource Memory}, and the \textbf{Knowledge Vault} (As shown in Figure \ref{fig:overall_structure}).
% Episodic, Semantic, and Procedural Memory follow prior formulations. 
Episodic Memory stores user-specific events and experiences; Semantic Memory captures concepts and named entities (e.g., the meaning of a new phrase or an understanding of a person); Procedural Memory records step-by-step instructions for performing tasks. Resource Memory is designed to store documents, files, and other media shared by the user. Knowledge Vault holds critical verbatim information that must be preserved exactly, such as addresses, phone numbers, email accounts, and other sensitive facts.

Each memory component is internally organized using a hierarchical structure. For example, Episodic Memory includes fields such as \texttt{summary} and \texttt{details}, while Semantic Memory organizes information by \texttt{name} and \texttt{description}. Managing this structured and heterogeneous memory is challenging for a single agent. Therefore, we adopt a multi-agent architecture: six Memory Managers for six memory components, and a \textbf{Meta Memory Manager} responsible for task routing. While this memory system can be plugged and connected with other existing agents, we build an extra \textbf{Chat Agent} to demonstrate how we can interact with an agent that has access to our memories. This memory system, which we call \textbf{\ours}, is modular and designed as a comprehensive and full memory system for LLM-based agents. Moreover,
when interacting with the Chat Agent, we propose an Active Retrieval mechanism, where the agent is required to generate a \emph{topic} before answering the question or executing the next step, and the retrieved information is inputted into the model in the system prompt. Meanwhile, we design multiple retrieval tools so that the agent can choose appropriate ones in response to different situations.

We evaluate \ours in two experimental settings. First, we introduce a challenging benchmark that requires extracting information and building memory from multimodal input. To this end, we collect between 5,000 and 20,000 high-resolution screenshots spanning one month of computer usage from three PhD students and construct evaluation questions grounded in their visual activity history. This setting demands sophisticated memory modeling and exceeds the capacity of existing long-context models. For example, each screenshot ranges from 2K to 4K resolution depending on the user's monitor, which limits Gemini to processing no more than 500 images at full resolution or approximately 3,600 images when resized to 256×256 pixels. Second, following Mem0~\citep{mem0}, we assess performance on the LOCOMO dataset~\citep{locomo}, which comprises long-form, multi-turn conversations. While each conversation is relatively short (around 26,000 tokens on average), we constrain the Chat Agent to answer questions using only retrieved memories, without access to the original conversation transcripts. This setup enables us to evaluate whether our system can effectively distill and route essential information into memory. In terms of results, because no existing memory systems can handle such a large volume of multimodal input, we compare against retrieval-augmented generation (RAG) baselines and long-context baselines. \ours achieves a 35\% improvement over RAG baselines while reducing storage requirements by 99.9\%, and a 410\% improvement over long-context baselines with a 93.3\% reduction in storage. On LOCOMO, \ours reaches state-of-the-art performance with an overall accuracy of 85.38\%, outperforming the best existing method by 8.0\% and approaching the upper bound set by long-context models. 
To make our memory system more accessible, we also developed a personal assistant powered by \ours. With the user’s permission to capture screen content, the assistant continuously builds a memory from screenshots and can answer any questions related to this accumulated information. This represents our effort to bring advanced memory capabilities to everyone, enabling users to experience the benefits of our system firsthand.

Our contributions are summarized as follows:
\begin{itemize}[leftmargin=*]
\item We analyze the limitations of existing memory architectures for LLM-based agents and propose a novel memory system composed of six specialized components and eight different agents. 
\item We introduce a new benchmark in which the agent must interpret a large collection of screenshots to build an understanding of user behavior—posing a significant challenge to the memory.
\item Through experiments on our new benchmark and existing benchmark LOCOMO, we demonstrate that \ours significantly outperform existing memory systems.
\item To make our approach accessible, we build and release a personal assistant application powered by \ours, enabling users to experience advanced memory capabilities in real-world scenarios.
\end{itemize}

%% file: 9_usecase.tex
\section{Application \& Use Cases}

\subsection{\ours Application}

To demonstrate the full functionality of our memory system, we developed a cross-platform application using \texttt{React-Electron} for the frontend and \texttt{Uvicorn} as the backend server. Moreover, we release the obtained file for direct installation. Within the application, users can activate screen monitoring by selecting the \texttt{ScreenShots} tab. This enables the agent to observe the user’s screen activity and dynamically update its memory, gradually building a contextual understanding of the user over time. With the obtained memory, the agent can answer questions related to the memories.

\paragraph{Memory Updates}
The application captures a screenshot every 1.5 seconds. To reduce redundancy, images that are visually similar to previously captured ones are discarded. Once 20 unique screenshots are collected, the memory update process is triggered—typically around every 60 seconds. During this process, relevant information is extracted and incorporated into the system's memory components. To reduce latency in processing user screenshots, we adopt a streaming upload strategy. Instead of batching and sending 20 images at once, we upload each screenshot immediately upon receiving it from the front-end. By leveraging the Gemini API—which supports loading images via Google Cloud URLs—we can efficiently transmit visual data without waiting for the full batch to accumulate. This approach significantly reduces end-to-end latency from approximately 50 seconds (as observed when using GPT-4 with direct image upload) to under 5 seconds using Gemini.

\begin{figure}[t]
    \centering
    \includegraphics[width=0.95\linewidth]{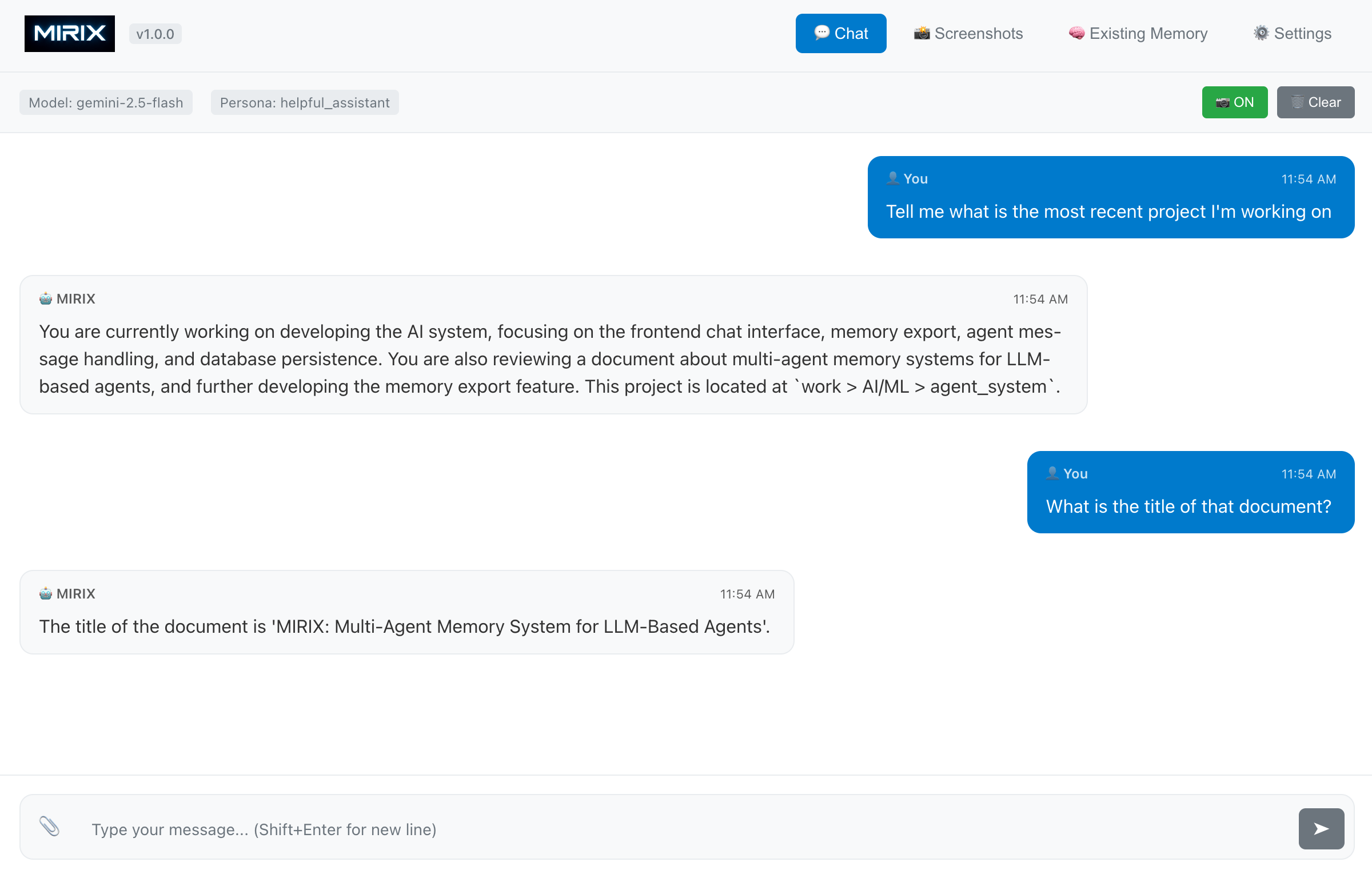}
    \caption{Chat Window}
    \label{fig:chat_window}
\end{figure}

\paragraph{Chat Interface}
The chat interface allows users to interact with the agent, which has full access to its accumulated memories. As illustrated in Figure~\ref{fig:chat_window}, users can query the assistant about past activities, and the agent can respond based on its memory content, enabling more informed and personalized interactions.

\begin{figure}
    \centering
    \includegraphics[width=1.0\linewidth]{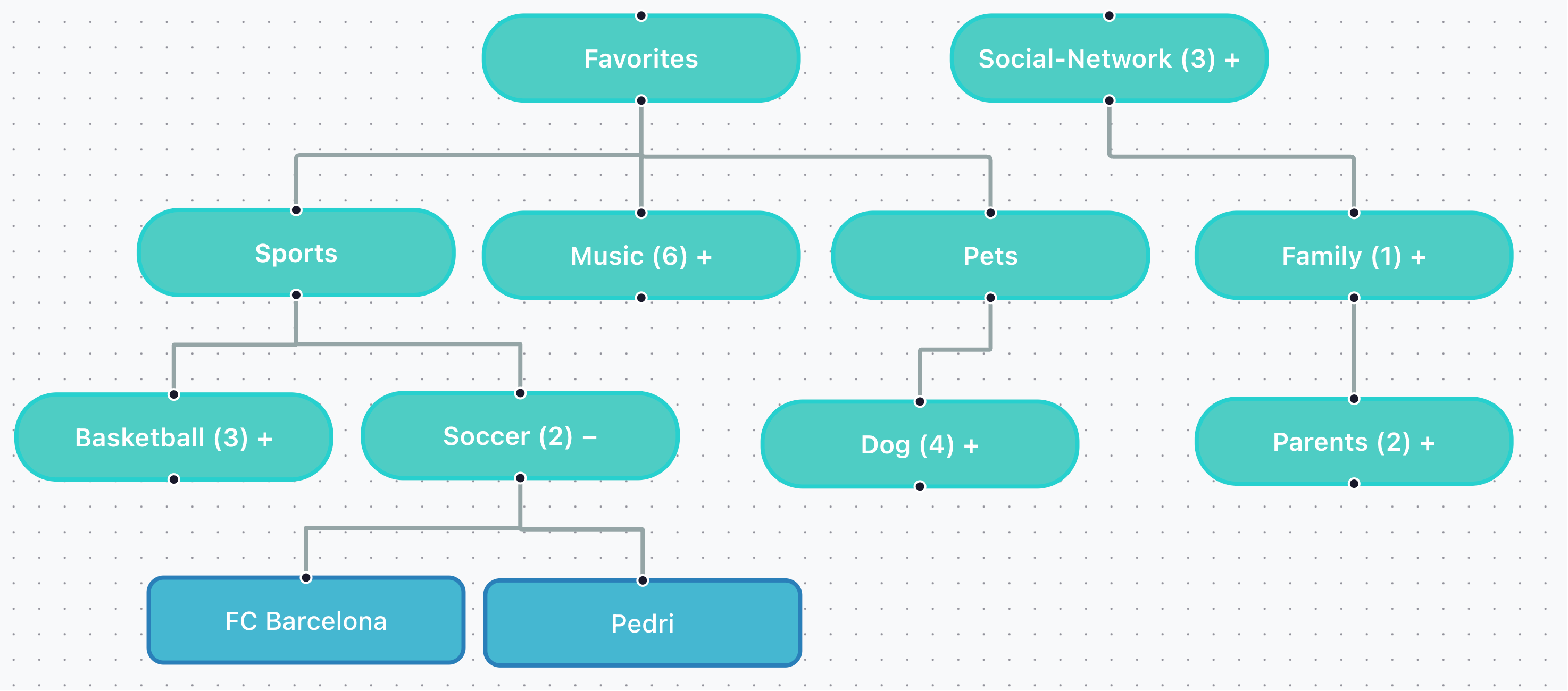}
    \caption{Tree Structure of an Example Semantic Memory. In this example, the user's Semantic Memory, which stores new concepts and their relationships, is organized hierarchically into multiple categories, such as \emph{Social Network} and \emph{Favorites}. Within \emph{Favorites}, the memory is further divided into more specific classes, including \emph{Sports}, \emph{Pets} and \emph{Music}.}
    \label{fig:semantic_tree}
\end{figure}

\begin{figure}
    \centering
    \includegraphics[width=1.0\linewidth]{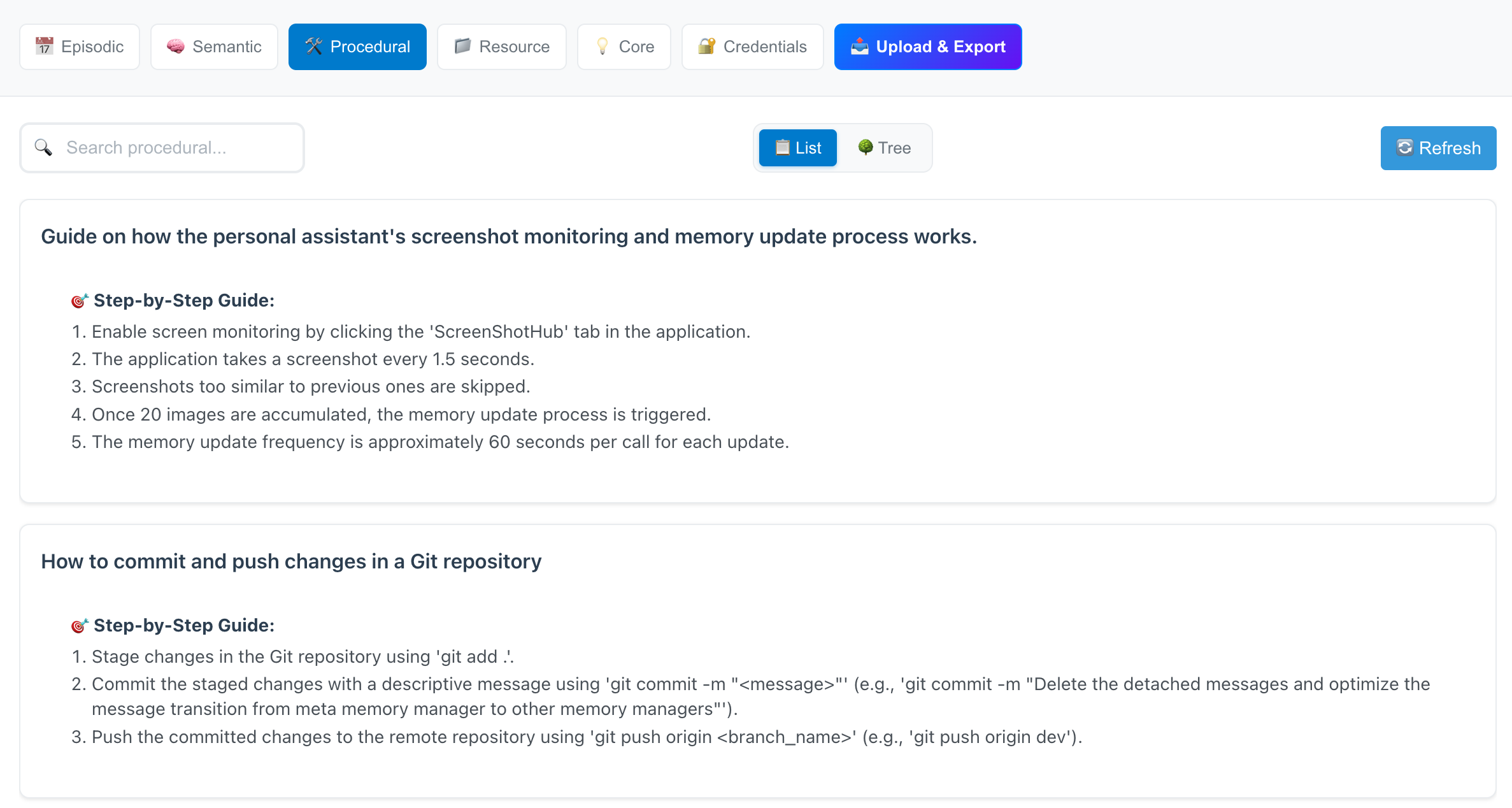}
    \caption{List View of an Example Procedural Memory.}
    \label{fig:procedural_list}
\end{figure}

\paragraph{Memory Visualization}
After observing the screen for a sufficient period, the agent organizes its knowledge into structured memory components. An example of Semantic Memory (organized into a tree structure) is shown in Figure~\ref{fig:semantic_tree}. We also provide a list view of the memories. An example of Procedural Memory is shown in Figure \ref{fig:procedural_list}.

\subsection{Memory System for Wearable Devices}
The wearable device market has seen rapid growth in recent years, driven by the increasing demand for intelligent, always-available personal assistants. Products like AI-powered glasses (e.g., Meta Ray-Ban, XREAL Air) and AI pins (e.g., Humane, Rabbit R1) aim to integrate seamless interaction into daily life through voice commands, visual capture, and real-time feedback. However, these devices often lack a long-term memory component that allows them to evolve with the user—retaining useful information over time, adapting to personal routines, and referencing past interactions in context-aware ways.

Our memory system is well-suited for integration into such wearable devices. By continuously collecting and processing data streams such as audio, visual scenes, and user queries, our system enables real-time memory formation. For instance, AI glasses equipped with our memory systems can automatically summarize meetings, remember frequently visited places, recognize recurring visual patterns, and recall previous conversations or tasks. With \ours, it can also evolve with the user and build a memory that is specifically made for the user. 
Moreover, the system’s modular memory architecture—including procedural, episodic, semantic, and resource memory—aligns naturally with the needs of lightweight, on-the-go devices. Procedural memory enables the assistant to learn user habits (e.g., daily routes, meeting structures), while semantic memory stores general knowledge about the user's preferences, environment, and routines. Episodic memory captures time-stamped, situational experiences, and can be queried for recalling specific events (e.g., “What did I see at the conference last week?”). Semantic Memory, on the other hand, can help organize the clients someone has seen over the past week and list them in a tree structure with details of their discussions.

Given the constraints of wearable hardware (limited compute and storage), our design also supports hybrid on-device/cloud memory management. Critical information in Knowledge Vault can be stored locally while large-scale memories such as Resource Memory can be offloaded and retrieved from the cloud on demand. 

In summary, our memory system serves as a cognitive backbone for wearable AI agents—enabling personalization, continuity, and intelligence at the edge. As the wearable market matures, embedding persistent, structured memory will be a key differentiator for next-generation AI assistants.

\subsection{Agent Memory Marketplace}
We envision a future where personal memory—collected and structured through AI agents—becomes a new digital asset class. In the AI era, memory is no longer just a passive log of past events, but an active, evolving knowledge base that can be shared, personalized, and monetized. The Agent Memory Marketplace is our proposal for a decentralized ecosystem where memory is exchanged, reused, and built collaboratively through AI agents.

We begin with a core belief: \textbf{human memory will become the most valuable and irreplaceable asset in the age of AI}. Unlike static data, memory encompasses lived experiences, subjective context, preferences, and interactions—making it deeply personal, yet highly reusable by intelligent systems. the marketplace is structured into three key layers:

\begin{enumerate}[leftmargin=*]
    \item \textbf{AI Agents Infrastructure}. Our technology provides infrastructure for lifetime intelligent, interactive agents. Examples include (1) Personal AI Assistants \& Companions: Tailored agents that continuously learn and evolve with the user. (2) AI Wearables: AI agents embedded in decentralized physical infrastructure (e.g., smart glasses, AI pins) that extend memory capture to the real world. (3) Multi-Agent Systems: Collaborative agents with shared memory access, enabling coordination and collective intelligence.
    \item \textbf{Privacy-Preserving Memory Infrastructure} To support trust and adoption, we aim to build in a robust privacy layer: (1) Encryption Layer: All memories are stored using end-to-end encryption. (2) Privacy Control: Fine-grained permissions allow users to choose which parts of their memory to share, trade, or restrict. (3) Decentralized Storage: Memories are stored in a distributed, censorship-resistant infrastructure.
    \item \textbf{Memory Marketplace and Social Function}: A peer-to-peer ecosystem for sharing, aggregating, and trading memories encoded in AI agents. Use cases include: (1) Memory Social/Trading: Tokenized exchange of memories (e.g., productivity hacks, niche workflows, or life advice). (2) Expert Communities: Collective memory-building for domain-specific expertise such as finance, education, or pet care. (3) Fan Economy and Dating Applications: Users can subscribe to AI personas based on influencer or celebrity memories, allowing fans to interact with memory-rich digital replicas of their favorite personalities. This technology also enables the creation of AI clones for accelerated dating and matching processes, where users can engage with AI agents in interactions before meeting in person.
\end{enumerate}

In summary, we envision a future where personal memory transcends its traditional role to become an active, valuable digital asset. By combining lifelong AI agents, privacy-preserving infrastructure, and a decentralized marketplace, we aim to create an ecosystem where memories can be securely captured, meaningfully shared, and collectively advanced. This approach not only empowers individuals to benefit from their own experiences but also unlocks new possibilities for collaboration, personalization, and economic value in the AI era.

%% file: 3_method.tex
\section{Methodology}

\subsection{Memory Components}

We design a modular memory architecture consisting of six distinct components: \textbf{Core Memory}, \textbf{Episodic Memory}, \textbf{Semantic Memory}, \textbf{Procedural Memory}, \textbf{Resource Memory}, and the \textbf{Knowledge Vault}. Each component is structurally and functionally tailored to capture different aspects of user interaction and world knowledge, enabling the agent to retrieve, reason, and act effectively across time and tasks.

\paragraph{Core Memory} 
Core Memory stores high-priority, persistent information that should always remain visible to the agent when engaging with the user. Inspired by the design of MemGPT~\citep{memgpt}, this memory is divided into two primary blocks: \texttt{persona} and \texttt{human}. The \texttt{persona} block encodes the identity, tone, or behavior profile of the agent, while the \texttt{human} block stores enduring facts about the user such as their name, preferences, and self-identifying attributes (e.g., ``User’s name is David'', ``User enjoys Japanese cuisine''). When the memory size exceeds 90\% of capacity, the system triggers a controlled rewrite process to maintain compactness without losing critical information. 

\paragraph{Episodic Memory} 
Episodic Memory captures time-stamped events and temporally grounded interactions that reflect the user's behavior, experiences, or activities. It functions as a structured log or calendar, enabling the agent to reason about user routines, recency, and context-aware follow-ups~\citep{Liao2024,episodic_memory_missing_piece,echo}. Each entry is defined by the following fields: \texttt{event\_type} (e.g., \texttt{user\_message}, \texttt{inferred\_result}, \texttt{system\_notification}), \texttt{summary} (a concise natural language description of the event), \texttt{details} (extended contextual information, including dialog excerpts or inferred states), \texttt{actor} (the origin of the event, either \texttt{user} or \texttt{assistant}), and \texttt{timestamp} (e.g., ``2025-03-05 10:15''). This structure allows the agent to temporally index memory and track change over time, such as identifying ongoing tasks or following up on pending actions.

\paragraph{Semantic Memory} 
Semantic Memory maintains abstract knowledge and factual information that is independent of specific times or events. This component serves as a knowledge base for general concepts, entities, and relationships — whether about the world or the user's social graph. For example, it may store entries like ``Harry Potter is written by J.K. Rowling'' or ``John is a friend of the user who enjoys jogging and lives in San Francisco.'' Each entry includes a \texttt{name} (the concept or entity identifier), \texttt{summary} (a concise definition or relationship statement), \texttt{details} (expanded background or contextual explanation), and \texttt{source} (e.g., \texttt{user\_provided}, \texttt{Wikipedia}, or inferred from conversation). Unlike episodic memory, semantic entries are intended to persist unless conceptually overwritten and support reasoning over social, geographic, or commonsense knowledge.

\begin{figure}
    \centering
    \includegraphics[width=0.8\linewidth]{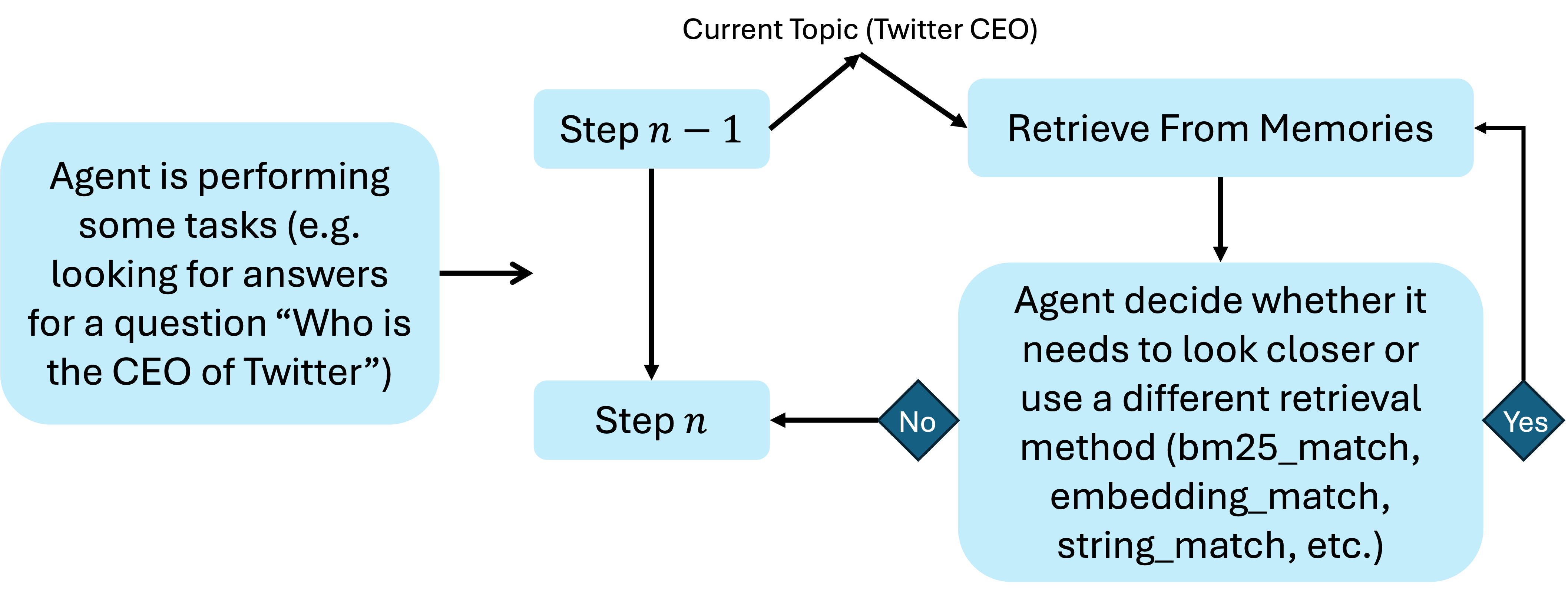}
    \caption{Demonstration of Active Retrieval.}
    \label{fig:active_retrieval}
\end{figure}

\paragraph{Procedural Memory} 
Procedural Memory stores structured, goal-directed processes such as how-to guides, operational workflows, and interactive scripts. These are neither time-sensitive (episodic) nor abstract facts (semantic) but instead represent actionable knowledge that can be invoked to assist the user with complex tasks. Typical examples include ``how to file a travel reimbursement form,'' ``steps to set up a Zoom meeting,'' or ``how to book a restaurant via OpenTable.'' Each entry includes an \texttt{entry\_type} (\texttt{workflow}, \texttt{guide}, or \texttt{script}), a \texttt{description} of the goal or function, and \texttt{steps} expressed as a list of instructions (optionally in JSON or structured format). This memory component supports instructional planning, automation, and decomposition of user goals into sub-tasks.

\paragraph{Resource Memory} 
Resource Memory handles full or partial documents, transcripts, or multi-modal files that the user is actively engaged with but do not fit into other memory categories. For instance, if the user is reading a friend's detailed picnic plan or a project proposal document, the agent can store and retrieve that information from Resource Memory. This component enables context continuity in long-running tasks. Each entry includes a \texttt{title} (resource name), \texttt{summary} (brief overview and context), \texttt{resource\_type} (e.g., \texttt{doc}, \texttt{markdown}, \texttt{pdf\_text}, \texttt{image}, \texttt{voice\_transcript}), and the full or excerpted \texttt{content}. This design enables the agent to reference previously seen material, quote from documents, or search within them to aid the user’s workflow.

\paragraph{Knowledge Vault} 
The Knowledge Vault serves as a secure repository for verbatim and sensitive information such as credentials, addresses, contact information, and API keys. These entries are not typically relevant to conversation-level reasoning but are crucial for performing authenticated tasks or storing long-term identifiers. Each entry includes an \texttt{entry\_type} (e.g., \texttt{credential}, \texttt{bookmark}, \texttt{contact\_info}, \texttt{api\_key}), a \texttt{source} (e.g., \texttt{user\_provided}, \texttt{github}), a \texttt{sensitivity} level (\texttt{low}, \texttt{medium}, \texttt{high}), and the actual \texttt{secret\_value}. Entries with high sensitivity are protected via access control and excluded from casual retrieval to prevent misuse or leakage.

\subsection{Active Retrieval and Retrieval Design}
In many memory-augmented systems (e.g., Mem0~\citep{mem0}, MemGPT~\citep{memgpt}), memory retrieval must be explicitly triggered. Otherwise, the language model often defaults to its parametric knowledge, which may be outdated or incorrect. For example, suppose the user previously said, “The CEO of Twitter is Linda Yaccarino,” and this information was saved in memory. A few days later, when this message is no longer present in the conversation history, the user might ask, “Who is the CEO of Twitter?” In this case, the language model may rely on outdated knowledge and incorrectly respond with “Elon Musk.” While explicitly instructing the model to “search your memory” can mitigate such errors, doing so repeatedly is impractical in natural conversations.

\begin{figure}
    \centering
    \includegraphics[width=\linewidth]{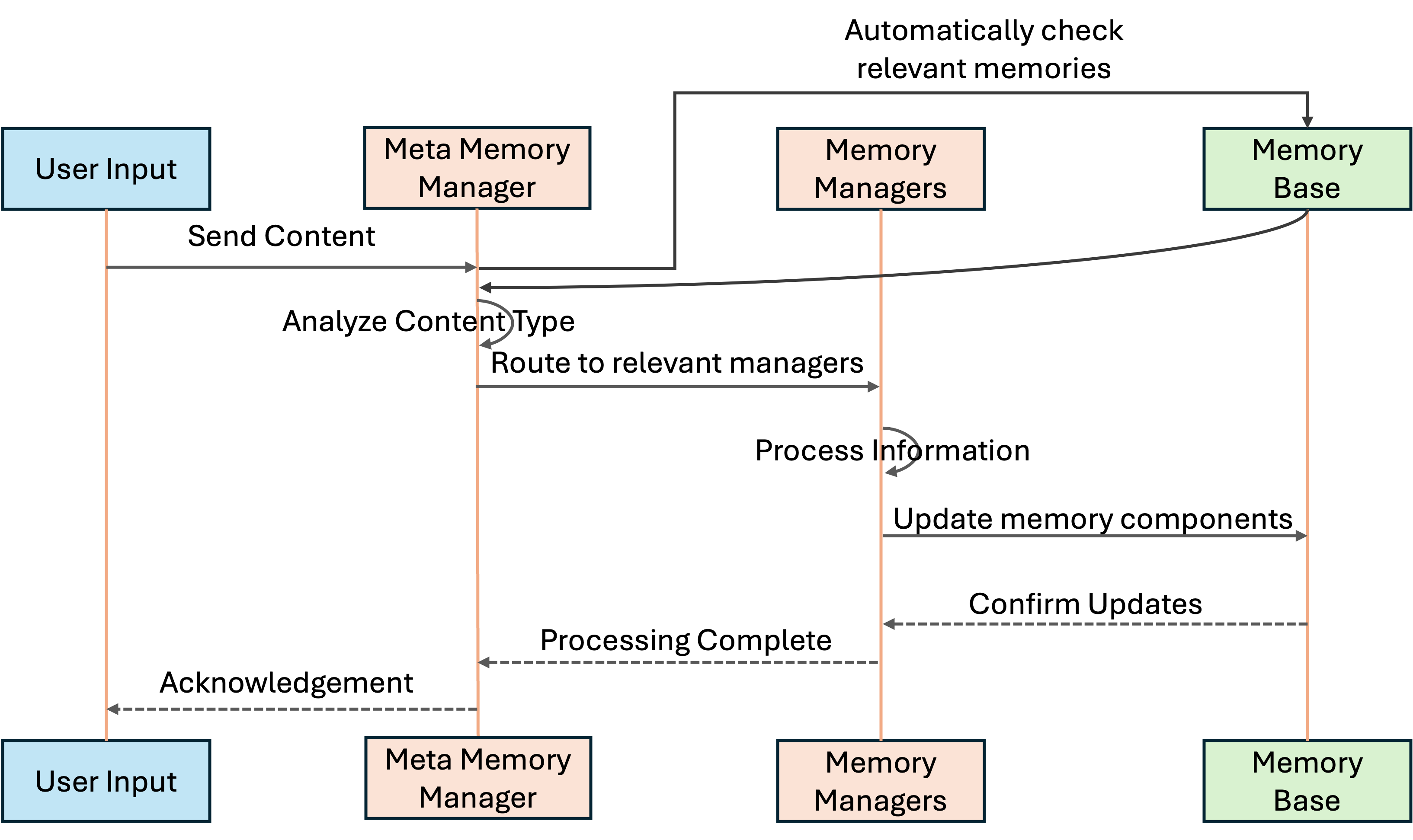}
    \caption{The Workflow of Memory Update.}
    \label{fig:memory_update_workflow}
\end{figure}

To address this, we propose an \textbf{Active Retrieval} mechanism. As illustrated in Figure~\ref{fig:active_retrieval}, the system operates in two stages: first, the agent generates a \emph{current topic} based on the input context; second, this topic is used to retrieve relevant memories from each memory component. Retrieved results are then injected into the system prompt. For instance, given the query “Who is the CEO of Twitter?”, the agent may infer the topic “CEO of Twitter”, which is then used to retrieve the top-10 most relevant entries from each of the six memory components. Retrieved content is tagged according to its source, such as \texttt{<episodic\_memory>}... \texttt{</episodic\_memory>}, ensuring the model is aware of both the content and its origin. This automatic retrieval pipeline eliminates the need for explicit user prompts to trigger memory access and ensures the model can incorporate up-to-date, personalized, or contextual information during response generation.

In addition to Active Retrieval, we support multiple retrieval functions, including \texttt{embedding\_match}, \texttt{bm25\_match}, and \texttt{string\_match}. We are actively expanding this set with more diverse and specialized retrieval strategies, ensuring that each method is well-differentiated so the agent can choose the most appropriate one to invoke based on context.

\subsection{Multi-Agent Workflow}

To manage the dynamic and heterogeneous nature of user interactions, we adopt a modular multi-agent architecture. This system orchestrates input processing, memory updating, and information retrieval across six distinct memory components through a coordinated and efficient workflow. The overall system is governed by a central \textit{Meta Memory Manager} and a set of specialized \textit{Memory Managers}, each responsible for maintaining one memory type.

\paragraph{Memory Update Workflow}
As illustrated in Figure~\ref{fig:memory_update_workflow}, when new input is received from the user, the system first automatically performs a search over the memory base. The retrieved information, together with the user input, is passed to the \emph{Meta Memory Manager}. The \emph{Meta Memory Manager} then analyzes the content and determines which memory components are relevant, routing the input to the corresponding \emph{Memory Managers}. These \emph{Memory Managers} update their respective memories in parallel while ensuring that redundant information is avoided within each memory type. After completing the updates, they report back to the \emph{Meta Memory Manager}, which finally sends an acknowledgment confirming that the memory update process is complete.

\paragraph{Conversational Retrieval Workflow}
For interactive dialogues, the \textit{Chat Agent} manages natural language communication with the user. To ground its responses in prior knowledge, it first performs an automatic search over the memory base upon receiving a user query. This initial search is a coarse retrieval spanning all six memory components and returns high-level summaries rather than detailed content. The Chat Agent then analyzes the query to determine which memory components warrant more targeted searches and selects appropriate retrieval methods accordingly. After obtaining the relevant results, it consolidates the information and synthesizes the final response. Moreover, if the user’s query involves updating memory—for example, providing new facts or corrections-the Chat Agent can interact directly with the corresponding Memory Managers to apply precise updates to specific memory components.

%% file: 4_experiments.tex
\begin{figure}
    \centering
    \includegraphics[width=\linewidth]{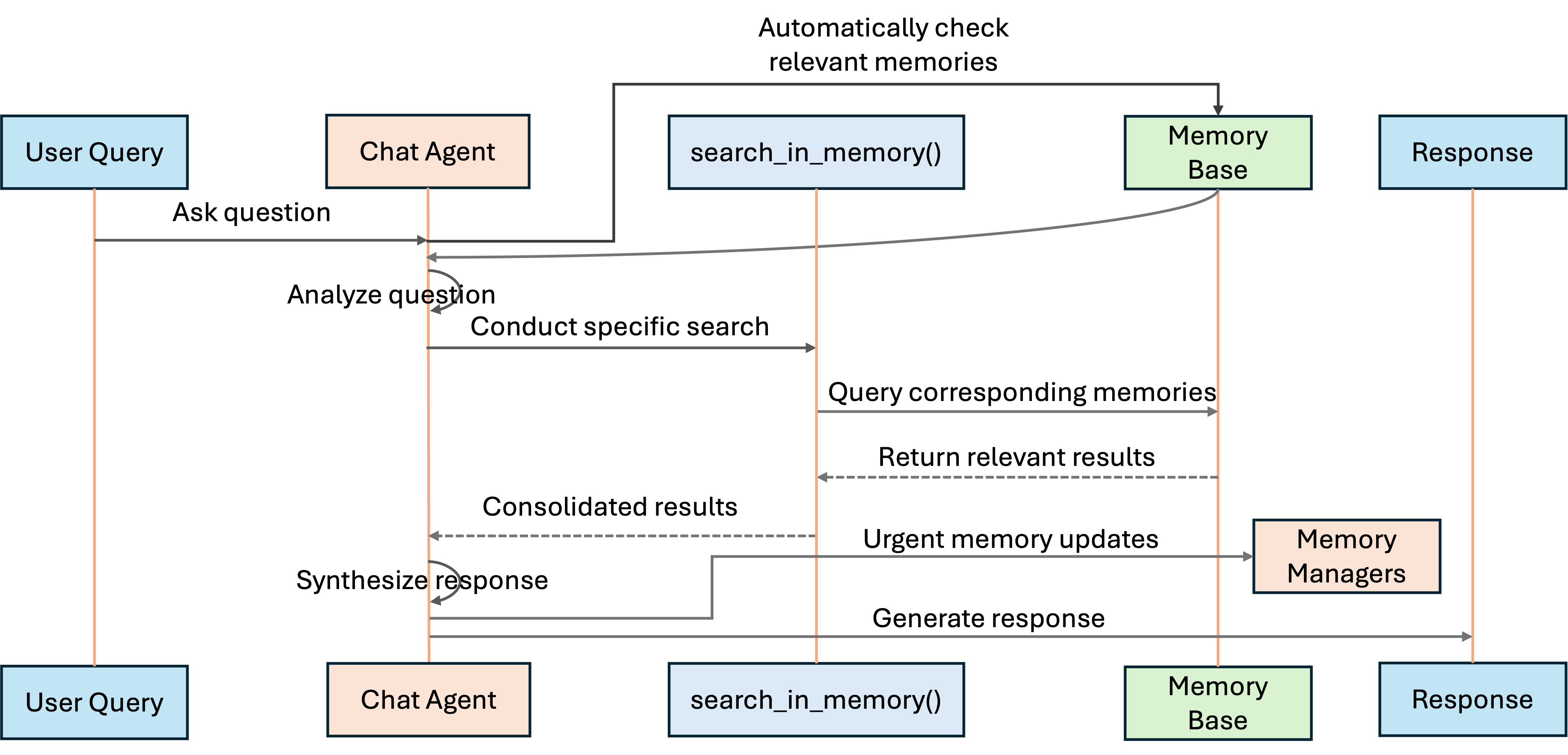}
    \caption{The Workflow of Responding to the User's Query.}
    \label{fig:memory_update_workflow}
\end{figure}

\section{Experiments}

\subsection{Experimental Setup}

\subsubsection{Datasets}
\paragraph{ScreenshotVQA}
We collect a new dataset that contains three PhD students' activities. We created a script that takes a screenshot every second. If the image taken this second is too similar to the last one (similarity $>$ 0.99) then we skip the current image. With this script, three PhD students majored in Computer Science and Physics run the program from a week to a month. The first student is a heavy computer user, which leads to 5,886 images in one day (03/09/2025). The second student uses the computer slightly lighter, leading to 18,178 images for 20 days (05/16/2025-06/06/2025). The third student is the lightest computer user, having 5349 images for over a month (05/02/2025-06/14/2025). Then, with these screenshots, we ask each student to manually create some questions and then we double-check the questions to make sure they are answerable. Eventually, we have 11 questions from the first user, 21 questions from the second user, and 55 questions from the third user. 

\paragraph{LOCOMO} Following Mem0~\citep{mem0}, we choose LOCOMO dataset for a vertical comparison between \ours and existing memory systems. LOCOMO has 10 conversations, where each conversation has 600 dialogues and 26000 tokens on average. There are averagely 200 questions for each conversation, which is suitable for the evaluation for memory systems where we are reuiqred to inject the conversation into the memory and then answer the 200 questions according to the obtained memory. The questions can be classified into multiple categories: single-hop, multi-hop, temporal and open-domain. The dataset also has another category called "adversarial", which is used to test whether the system can identify unanswerable questions. Following Mem0~\citep{mem0}, we exclude this category in our setting to provide fair comparisons with the earlier methods.

\subsubsection{Evaluation Metrics and Implementation Details}
\paragraph{Evaluation Metrics}
In this paper, for both of the above datasets, we mainly consider the metric \textbf{LLM-as-a-Judge}. Specifically, we use \texttt{GPT-4.1} as the judge to look at the question, answer, and response to predict whether the response addresses the question successfully. 

\paragraph{Implementation Details for ScreenshotVQA}
Across all experiments, we use the \texttt{gemini-2.5-flash-preview-04-17} model as the backbone. We selected Gemini because it integrates seamlessly with Google Cloud, enabling asynchronous image uploads and retrieval. This significantly accelerates processing, as each step requires multiple function calls: one to the meta memory manager and between zero and six to the other memory managers.

\paragraph{Implementation Details for LOCOMO} 
For \ours, since every agent needs to call many functions to successfully insert the information into the memory systems, it requires the language model to have strong abilities on function calling. To this end, we use \texttt{gpt-4.1-mini} as our backbone model as it shows a stronger ability than \texttt{gpt-4o-mini} in terms of function calling. This is also revealled in Berkeley Function Calling Benchmark~\citep{berkeley-function-calling-leaderboard} where \texttt{gpt-4o-mini} has multi-turn overall acc as 22.12, vs the acc of \texttt{gpt-4.1-mini} being 29.75. For LOCOMO dataset, we run the baselines LangMem, RAG-500, Mem0 using the code provided in Mem0~\footnote{\url{https://github.com/mem0ai/mem0/tree/main/evaluation}} with setting \texttt{MODEL} in the \texttt{.env} file as \texttt{gpt-4.1-mini}. Then for Zep, we use the code from their official repo\footnote{\url{https://github.com/getzep/zep-papers?ref=www.getzep.com/blog}} and replace the backbone model with \texttt{gpt-4.1-mini}. For all the baselines, we run their code once, while for \ours and Full-Context method, we run three times to report the average scores. We provide the results of each single run in Appendix \ref{sec:full_experimental_results}. 
The complete code for our evaluation and the predicted results from various baselines and \ours are provided in the \texttt{public\_evaluation} branch\footnote{\url{https://github.com/Mirix-AI/MIRIX/tree/public_evaluation}} in our official repository.

\subsection{Experimental Results on ScreenshotVQA}
For \textbf{ScreenshotVQA}, since existing memory systems such as Letta, Mem0 still lack the ability to process multimodal input, we omit the comparisons with them. We consider the following baselines: 

\textbf{Gemini}. As a long-context baseline, Gemini directly ingests the full set of screenshots to answer the questions. Because the original high-resolution screenshots exceed the model’s context window, we resize them to 256×256 pixels, enabling approximately 3,600 images to fit into a single prompt. We then include the most recent 3,600 images in the prompt and query the model for an answer. Specifically, we use the \texttt{gemini-2.5-flash-preview-04-17} model.

\textbf{SigLIP}. SigLIP serves as a retrieval-augmented baseline. We first use SigLIP to identify the top 50 most relevant images for each query and then prompt Gemini to generate the answer based on these retrieved images. In this setup, we employ \texttt{google/siglip-so400m-patch14-384} as the retriever and \texttt{gemini-2.5-flash-preview-04-17} as the language model.

We evaluate all methods using two metrics: (1) \textbf{Accuracy}, measured via an LLM-as-a-Judge approach comparing the generated responses to ground-truth answers and reporting the average accuracy across the three participants; and (2) \textbf{Storage}, defined as follows—For Gemini, we report the total size of the images after resizing to 256×256. For SigLIP, we report the total size of the original retrieved images in their native 2K–4K resolution. For \ours, we use \texttt{sqlite} as the storage backend and report the size of the \texttt{sqlite.db} file containing all extracted information.

The results are summarized in Table~\ref{tab:screenshotvqa_results}. As shown, \ours significantly outperforms all existing baselines while requiring substantially less storage. Specifically, because SigLIP retains the full set of high-resolution images, its storage footprint is very large, corresponding to the total size of all images in their original 2K–4K resolution. The long-context baseline (Gemini) reduces this by resizing images to 256×256 pixels, but still incurs considerable storage overhead. In contrast, \ours avoids storing raw images altogether and instead maintains only the compact \texttt{sqlite} database of extracted information, resulting in a much smaller storage size. Compared to retrieval-augmented generation (RAG) baselines, MIRIX achieves a 35\% improvement in accuracy while reducing storage requirements by 99.9\%. Relative to the long-context Gemini baseline, MIRIX yields a 410\% improvement in accuracy with a 93.3\% reduction in storage. 

\begin{table*}[]
    \centering
    \resizebox{\textwidth}{!}{
    \begin{tabular}{c|cc|cc|cc|cc}
    \toprule
        & \multicolumn{2}{c}{Student 1} &  \multicolumn{2}{c}{Student 2} &  \multicolumn{2}{c}{Student 3} & \multicolumn{2}{c}{Overall} \\
         & Acc $\uparrow$ & Storage$\downarrow$ & Acc $\uparrow$ & Storage$\downarrow$ & Acc $\uparrow$ & Storage$\downarrow$ & Acc $\uparrow$ & Storage$\downarrow$  \\
         \midrule
        Gemini & 0.0000 & 142.10MB & 0.0952 & 438.86MB & 0.2545 & 129.14MB & 0.1166 & 236.70MB \\
        SigLIP@50 & 0.3636 & 22.55GB & 0.4138 & 19.88GB & 0.5455 & 2.82GB & 0.4410 & 15.07GB \\
        \ours & \textbf{0.5455} & \textbf{20.57MB} & \textbf{0.5667} & \textbf{19.83MB} & \textbf{0.6727} & \textbf{7.28MB} & \textbf{0.5950} & \textbf{15.89MB} \\
        \bottomrule
    \end{tabular}}
    \caption{Experimental Results on ScreenshotVQA.}
    \label{tab:screenshotvqa_results}
\end{table*}

\begin{table*}[!t]
\centering
% \resizebox{\linewidth}{!}{%
\begin{tabular}{p{0.5cm}l|cccc|c}   % ← added a 1.5 cm column on the left
\toprule
& \textbf{Method} & \textbf{Single Hop} & \textbf{Multi-Hop} & \textbf{Open Domain} & \textbf{Temporal} & \textbf{Overall} \\
\midrule
\multirow{7}{*}{\rotatebox[origin=c]{90}{\textbf{gpt-4o-mini}}}
  & A-Mem      & 39.79 & 18.85 & 54.05 & 49.91 & 48.38 \\
& LangMem     & 62.23 & 47.92 & 71.12 & 23.43 & 58.10 \\
& OpenAI      & 63.79 & 42.92 & 62.29 & 21.71 & 52.90 \\
& Mem0        & 67.13 & 51.15 & 72.93 & 55.51 & 66.88 \\
& Mem0$^g$       & 65.71 & 47.19 & \textbf{75.71} & 58.13 & 68.44 \\
& Memobase    & 63.83 & 52.08 & 71.82 & 80.37 & 70.91 \\
& Zep         & 74.11 & 66.04 & 67.71 & 79.76 & 75.14\\
% \cmidrule{2-7}
% & Full-Context & 80.14 & 46.89 & 58.33 & 90.49 & 77.51 \\
\midrule
\multirow{6}{*}{\rotatebox[origin=c]{90}{\textbf{gpt-4.1-mini}}}
 % & Zep & 48.23 & 17.76 & 39.58 & 62.43 & 49.09 \\
& LangMem & 74.47 & 61.06 & 67.71 & 86.92 & 78.05 \\
& RAG-500 & 37.94 & 37.69 & 48.96 & 61.83 & 51.62 \\
& Zep & 79.43 & 69.16 & 73.96 & 83.33 & 79.09 \\
& Mem0    & 62.41 & 57.32 & 44.79 & 66.47 & 62.47 \\
\cmidrule{2-7}
& \textbf{\ours} & \textbf{85.11} & \textbf{83.70} & 65.62 & \textbf{88.39} & \textbf{85.38} \\
\cmidrule{2-7}
& Full-Context & 88.53 & 77.70 & 71.88 & 92.70 & 87.52 \\
\bottomrule
\end{tabular}%}
\caption{LLM-as-a-Judge scores (\%, higher is better) for each question type in the \texttt{LOCOMO} dataset. As mentioned in Mem0~\citep{mem0}, as the average length in this dataset is only 9k, Full-Context is essentially the upper-bound. Thus recovering the performance of Full-Context shows the advancements of \ours. }
\label{tab:locomo}
\end{table*}

\subsection{Experimental Results on LOCOMO}
We compare \ours against the following baselines:

\textbf{A-Mem~\citep{amem}:} A memory system that builds Zettelkasten-style knowledge graphs from user-agent interactions, dynamically linking notes using embedding similarity and LLM reasoning.

\textbf{LangMem\footnote{\url{https://github.com/langchain-ai/langmem}}:} LangChain’s long-term memory module that extracts and stores salient facts from conversations for later retrieval through retrievers like FAISS or Chroma.

\textbf{Zep~\citep{zep}:} A commercial memory API that constructs a temporal knowledge graph (Graphiti) over user conversations and metadata, designed for fast semantic querying.

\textbf{Mem0~\citep{mem0}:} An open-source memory system that incrementally compresses and stores memory facts using LLM-based summarization, with an optional graph memory extension.

\textbf{Memobase\footnote{\url{https://github.com/memodb-io/memobase}}:} A profile-based memory module that tracks persistent user attributes and preferences to enable long-term personalization.

All baselines are re-implemented using the same backbone model (\texttt{gpt-4.1-mini}). We also report the results shown in Mem0~\citep{mem0} where the backbone model is \texttt{gpt-4o-mini}. The results are shown in Table~\ref{tab:locomo}. From the table, we observe the following:

\textbf{Overall:} \ours achieves the highest average J score, outperforming all baselines by a significant margin. It improves upon the strongest open-source competitor, LangMem, by over 8 points.

\textbf{Single-Hop and Temporal:} On fact lookup and temporal-ordering tasks, \ours shows significantly better performances than baselines, which validates the effectiveness of our hierarchical memory storage. 
We note that there is a minor gap between \ours and the Full-Context baseline on Single-Hop questions. Upon reviewing the results, we identified a major reason for this. In some cases, the questions are ambiguous. For example, consider the question \emph{``When is Melanie planning on going camping?''} In the conversation history, Melanie stated \emph{``We’re thinking about going camping next month''} in May, which would suggest \emph{June} as the answer. However, she later mentioned in October, \emph{``Absolutely! It really helps me reset and recharge. I love camping trips with my fam, ’cause nature brings such peace and serenity,''} referring to a more recent trip. Because \ours saves the consolidated event \emph{``On 19 October 2023, Melanie and her family went camping after their road trip,''} it tends to prioritize the confirmed occurrence over the earlier plan, leading to discrepancies when the question expects the planned date rather than the actual event. Other similar ambiguities contribute to the slightly lower performance of \ours on Single-Hop questions.

\textbf{Multi-Hop:} \ours demonstrates the largest gains in this category, outperforming all baselines by over 24 points. For example, in questions such as \emph{“Where did Caroline move from 4 years ago?”}, where the correct answer is Sweden, the supporting evidence is dispersed across multiple parts of the conversation. One part may state \emph{“Caroline moved from her hometown 4 years ago”}, while an earlier statement establishes that \emph{“Caroline’s hometown is Sweden”}. For multi-hop questions like these, \ours achieves better performance because it explicitly stores the consolidated event \emph{“Caroline moved from her hometown, Sweden, 4 years ago”}, removing the need to stitch together partial information at query time. In contrast, full-context methods must first retrieve the partial answer \emph{“hometown”} and then figure out separately that \emph{“hometown”}  refers to \emph{“Swedon”}. These multi-hop questions might be easy for reasoning models like OpenAI-O3, but for non-reasoning models such as \texttt{gpt-4.1-mini}, this additional reasoning step might fail, leading to slightly inferior performance compared with \ours.

\textbf{Open-Domain:} While \ours performs well, the margin between ours and the baselines is narrower. This category of questions usually asks the agent ``what if'' questions, requiring the agent to infer across longer terms. The gap between \ours and Full-Context method show the inherent limitation of RAG methods, which is the lack of global understanding. While \ours is no longer simple RAG, we still rely on RAG to retrieve important information in the memory, which might lead to the bottleneck of our agent in this category.

In summary, these results demonstrate that \ours delivers state-of-the-art performance on LOCOMO while remaining highly efficient and modular. Its component-specific memory management and intelligent routing are particularly effective for long-range multi-hop reasoning.

%% file: 2_related_work.tex
\section{Related Work}

\paragraph{Memory-Augmented Large Language Models}
A growing body of work focuses on building latent-space memory systems, as characterized in M+\citep{mplus}, where transformer architectures are modified to support memory augmentation. These memory components can reside in various latent forms, including model parameters\citep{self-param}, external memory matrices~\citep{larimar,infini-attention}, hidden states~\citep{RMT, he2024camelot, memoryllm, mplus}, soft prompts~\citep{memoryTransformer}, and key-value caches~\citep{MemoryBank,cartridges,snapkv,h2o}. While these approaches demonstrate promising results and continue to advance the frontier of latent-space memory, most require retraining the model~\citep{memoryllm, mplus, infini-attention}, making them incompatible with powerful closed-source models like GPT-4 or DeepSeek-R1. Additionally, methods based on key-value caching rely on preserving past keys and values, functioning more as long-context methods rather than true memory systems that support abstraction, consolidation, and reasoning over stored experiences~\citep{lscs}.

\paragraph{Memory-Augmented LLM Agents}
Token-level memory remains the dominant approach in current LLM agents~\citep{mplus}, where past conversational content is stored in raw text form within external databases. Notable examples include commercial systems such as Zep~\citep{zep}, Mem0~\citep{mem0}, and MemGPT~\citep{memgpt}. These agents perform well on long-term conversational benchmarks~\citep{locomo,longmemeval} and document-based retrieval tasks. However, they often fall short in real-world applications due to simplistic memory architectures. Most notably, the absence of modular memory components hinders effective memory routing and leads to inefficiencies in retrieval and usage.

\paragraph{Various Memory Types}
Cognitive science broadly categorizes memory into short-term (working) memory and long-term memory~\citep{sridhar2023cognitive}. In the context of LLMs, short-term memory is often mapped to the input context window~\citep{li2024memory}, while long-term memory becomes a catch-all category for any information outside the context~\citep{mplus}. To address this limitation, recent works have proposed finer-grained memory architectures. For example, \citet{episodic_memory_missing_piece} emphasizes the importance of episodic memory in LLM agents. Other systems incorporate both episodic and semantic memory~\citep{arigraph,memory_consciousness_and_llm,tulving1985memory}. Semantic memory has also been highlighted as critical for real-world reasoning and abstraction~\citep{procedural_memory_all_you_need,kim2023machine}. In addition, procedural memory, responsible for learned skills and routine tasks, has been identified as another crucial component~\citep{procedural_memory_all_you_need}. Despite these advancements, existing methods stop at identifying individual memory types, and they are not formed into a comprehensive memory system.

\paragraph{Multi-Agent Systems}
Rather than relying on a monolithic agent, recent advances explore multi-agent frameworks where specialized agents coordinate to accomplish complex tasks. Early systems like AutoGPT and BabyAGI~\citep{AutoGPT,BabyAGI} adopt an autonomous planning-execution loop while maintaining a shared memory log. More recent designs introduce role specialization: MetaGPT~\citep{Hong2023} mimics a software development team structure, and AgentVerse~\citep{Chen2024} assigns agents to specific roles such as planning or evaluation. Cognitive theories also support modularity in memory, particularly distinctions between episodic and semantic types, as emphasized in \citet{Liao2024}. \ours builds on these ideas by deploying eight specialized agents, each managing a distinct memory type (e.g., episodic, semantic, procedural), and coordinating to process multi-modal inputs effectively.

%% file: 5_conclusion.tex
\section{Conclusion and Future Work}
In this work, we introduce \ours, a novel memory architecture designed to enhance the long-term reasoning and personalization capabilities of LLM-based agents. Unlike existing memory systems that primarily rely on flat storage or limited memory types, \ours leverages a structured and compositional approach, incorporating six specialized memory components—Core, Episodic, Semantic, Procedural, Resource, and Knowledge Vault—managed by dedicated Memory Managers under the coordination of a Meta Memory Manager. To rigorously evaluate our system, we introduce a challenging multimodal benchmark based on high-resolution screenshots of real user activity, demonstrating that \ours achieves substantial gains in accuracy and storage efficiency compared to both retrieval-augmented generation and long-context baselines. Experiments on the LOCOMO benchmark confirm that \ours delivers state-of-the-art performance in long-form conversational settings. Finally, to make these capabilities accessible to a broader audience, we build and release a personal assistant application powered by \ours, allowing users to experience consistent, memory-enhanced interactions in everyday scenarios. We hope this work paves the way for more robust, scalable, and human-like memory systems for LLM-based agents. In the future, we aim to build more challenging real-world benchmarks to comprehensively evaluate our system and constantly improve \ours and the associated personal assistant application to deliver better experiences to the users.

%% file: 6_appendix.tex
\section{Full Experimental Results with Different Runs}
\label{sec:full_experimental_results}

We run \ours and Full-Context with \texttt{gpt-4.1-mini} three times and we report the full results in Table \ref{tab:full_locomo}. There are variations across different runs, while \ours consistently achieves state-of-the-art results. Full predicted results and LLM-Judge scores are provided in the folder \url{https://github.com/Mirix-AI/MIRIX/tree/public_evaluation/public_evaluations/evaluation_metrics}.

\begin{table*}[h!]
\centering
% \resizebox{\linewidth}{!}{%
\begin{tabular}{p{0.5cm}l|cccc|c}   % ← added a 1.5 cm column on the left
\toprule
& \textbf{Method} & \textbf{Single Hop} & \textbf{Multi-Hop} & \textbf{Open Domain} & \textbf{Temporal} & \textbf{Overall} \\
\midrule
\multirow{8}{*}{\rotatebox[origin=c]{90}{\textbf{gpt-4o-mini}}}
  & A-Mem      & 39.79 & 18.85 & 54.05 & 49.91 & 48.38 \\
& LangMem     & 62.23 & 47.92 & 71.12 & 23.43 & 58.10 \\
& OpenAI      & 63.79 & 42.92 & 62.29 & 21.71 & 52.90 \\
& Mem0        & 67.13 & 51.15 & 72.93 & 55.51 & 66.88 \\
& Mem0$^g$       & 65.71 & 47.19 & \textbf{75.71} & 58.13 & 68.44 \\
& Memobase    & 63.83 & 52.08 & 71.82 & 80.37 & 70.91 \\
& Zep         & 74.11 & 66.04 & 67.71 & 79.76 & 75.14\\
\cmidrule{2-7}
& Full-Context & 80.14 & 46.89 & 58.33 & 90.49 & 77.51 \\
\midrule
\multirow{9}{*}{\rotatebox[origin=c]{90}{\textbf{gpt-4.1-mini}}}
  & LangMem & 76.24 & 63.24 & 63.54 & 88.11 & 79.22 \\
& RAG-500 & 37.94 & 37.69 & 48.96 & 61.83 & 51.62 \\
& Mem0    & 62.41 & 57.32 & 44.79 & 66.47 & 62.47 \\
\cmidrule{2-7}
& \textbf{\ours-Run1}   & 85.46 & 80.06 & 64.58 & 87.16 & 83.98 \\
& \textbf{\ours-Run2}   & 85.46 & 85.36 & 64.58 & 91.32 & 87.34 \\
& \textbf{\ours-Run3}   & 84.40 & 85.67 & 67.71 & 86.68 & 84.82 \\
\cmidrule{2-7}
& Full-Context-Run1 & 88.30 & 71.74 & 72.92 & 92.98 & 86.43 \\
& Full-Context-Run2 & 88.29 & 81.31 & 71.88 & 92.50 & 88.13 \\
& Full-Context-Run3 & 89.01 & 80.06 & 70.83 & 92.63 & 88.00 \\
\bottomrule
\end{tabular}%}
\caption{LLM-as-a-Judge scores (\%, higher is better) for each question type in the \texttt{LOCOMO} dataset. As mentioned in Mem0~\citep{mem0}, as the average length in this dataset is only 9k, Full-Context is almost the upper-bound. Thus recovering the performance of Full-Context shows the advancements of \ours. For Zep, we only achieve 49.09 overall score with \texttt{gpt-4.1-mini} using the implementation from mem0, we are afraid there might be errors in their implementation, so we skip the results here.}
\label{tab:full_locomo}
\end{table*}